\documentclass[runningheads]{llncs}
\usepackage{graphicx}
\usepackage{comment}
\usepackage{amsmath,amssymb} 
\usepackage{color}
\usepackage[super]{nth}
\usepackage{colortbl}
\usepackage{multirow}

\usepackage{soul}
\usepackage{url}
\usepackage[hidelinks]{hyperref}
\usepackage[utf8]{inputenc}
\usepackage[small]{caption}
\usepackage{graphicx}
\usepackage{mathrsfs}
\usepackage{tabularx,lipsum,environ,amsmath,amssymb}
\usepackage{hyperref}

\usepackage[table,xcdraw]{xcolor}
\usepackage{amsmath}
\usepackage{algorithm}
\usepackage{algorithmic}
\usepackage{booktabs}
\usepackage{amsfonts}
\usepackage{indentfirst}
\usepackage{diagbox}
\usepackage{subcaption}
\usepackage{cite}
\usepackage{amsbsy}
\usepackage{pifont}
\definecolor{mypink3}{cmyk}{0, 0.7808, 0.4429, 0.1412/}
\definecolor{myblue3}{cmyk}{0, 0.1418, 0.9412, 0.3412/}
\definecolor{red}{RGB}{255, 0, 0}

\hypersetup{colorlinks, citecolor=blue}

\newcommand{\xmark}{\ding{55}}%
\newcommand{\figref}[1]{Fig.~\ref{#1}}
\newcommand{\tabref}[1]{Table.~\ref{#1}}

\newcommand{\eqnref}[1]{Eq.~(\ref{#1})}

\DeclareMathOperator*{\argmax}{arg\,max}

\usepackage[width=122mm,left=12mm,paperwidth=146mm,height=193mm,top=12mm,paperheight=217mm]{geometry}

\begin{document}
\pagestyle{headings}
\mainmatter
\title{Two-phase Pseudo Label Densification for Self-training based Domain Adaptation} 


%
\author{%
  Inkyu Shin\quad%
  Sanghyun Woo\quad%
  Fei Pan\quad%
  In So Kweon
  %
  }
\authorrunning{I. Shin et al.}
\titlerunning{Two-phase Pseudo Label Densification}
\institute{KAIST, South Korea. \\
\email{\{dlsrbgg33,shwoo93,feipan,iskweon77\}@kaist.ac.kr}
}

\maketitle

\begin{abstract}
Recently, deep self-training approaches emerged as a powerful solution to the unsupervised domain adaptation.
The self-training scheme involves iterative processing of target data; it generates target pseudo labels and retrains the network.
However, since only the confident predictions are taken as pseudo labels, existing self-training approaches inevitably produce sparse pseudo labels in practice.
We see this is critical because the resulting insufficient training-signals lead to a sub-optimal, error-prone model.
In order to tackle this problem, we propose a novel \textbf{T}wo-phase \textbf{P}seudo \textbf{L}abel \textbf{D}ensification framework, referred to as \textbf{TPLD}.
In the first phase, we use sliding window voting to propagate the confident predictions, utilizing intrinsic spatial-correlations in the images. In the second phase, we perform a confidence-based easy-hard classification.
For the easy samples, we now employ their full pseudo-labels.
For the hard ones, we instead adopt adversarial learning to enforce hard-to-easy feature alignment.
To ease the training process and avoid noisy predictions, we introduce the bootstrapping mechanism to the original self-training loss.
We show the proposed TPLD can be easily integrated into existing self-training based approaches and improves the performance significantly. Combined with the recently proposed CRST self-training framework, we achieve new state-of-the-art results on two standard UDA benchmarks.
\keywords{Unsupervised Domain Adaptataion, Self-training}
\end{abstract}

\section{Introduction}
Unsupervised domain adaptation (UDA) aims to transfer knowledge learned from the label-rich source domain to an unlabeled new target domain.
It is a practical and crucial problem as it could be beneficial for various label-scarce real-world scenarios, e.g., simulated learning for robots~\cite{sim2robot} or autonomous driving~\cite{Richter_2016_ECCV}. 
In this paper, we focus on the UDA for semantic segmentation, aiming to adopt a source segmentation model to a target domain without any labels.

The dominant paradigm in UDA is based on \textbf{\textit{adversarial learning}}~\cite{tsai2018learning, Hong_2018_CVPR, pmlr-v80-hoffman18a, Chen_2019_CVPR, Yawei2019Taking, vu2019advent}. In particular, it minimizes both (source domain) task-specific loss and domain adversarial loss. The method thus retains good performance on the source domain task, and at the same time, can bridge the gap between source and target feature distributions. 
While the adversarial learning has achieved great success in UDA, recently another line of studies using \textit{\textbf{self-training}} emerged~\cite{zou2018unsupervised, Zou_2019_ICCV}.
Self-training generates a set of pseudo labels corresponding to high prediction scores in the target domain and then re-trains the network based on the generated pseudo labels.
Recently, Zou \& Yu have proposed two seminal works on CNN-based self-training methods; class balanced self-training (CBST)~\cite{zou2018unsupervised}, and confidence regularized self-training (CRST)~\cite{Zou_2019_ICCV}.
Unlike adversarial learning methods which utilize two separate losses, CBST presents a single unified self-training loss. 
It allows learning of domain-invariant features and classifiers in an end-to-end manner, both from labeled source data and pseudo labeled target data. 
CRST further generalizes the feasible space of pseudo labels and adopts regularizer.
These self-training methods show state-of-the-art results in multiple UDA settings. 
However, we observe that its internal pseudo label selection tends to excessively cut-out the predictions, which often leads to sparse pseudo labels. We argue that sparse pseudo labels significantly miss meaningful training signals, and thus, the final model may deviate from the optimal solution eventually.
A natural way to obtain dense pseudo labels is by lowering the selection threshold.
However, we observe this naive approach brings noisy, unconfident predictions at an early stage, and this accumulates and propagates the errors.

To effectively address this issue, we present a two-step, gradual pseudo label densification method.
The overview is shown in~\figref{fig:architecture}.
In the first phase, we use sliding window voting to propagate the confident predictions, utilizing the intrinsic spatial correlations in the images.
In the second phase, we perform an easy-hard classification using a proposed image-level confidence score.
Our intuition is simple: As the model improves over time, its predictions can be trusted more. 
Thus, if the model in the second stage is confident with their prediction, we now do not zero out them.
Indeed, we empirically observe that the confident, easy samples are near to the ground truth and vice versa.
This motivates us to utilize full pseudo labels for the easy samples, while for the hard samples, we enforce adversarial loss to learn hard-to-easy adaption.
Meanwhile, to tackle noisy labels effectively for both first and second phase training, we introduce the bootstrapping mechanism into the self-training loss function.
By connecting all together, we build a two-phase pseudo label densification framework called TPLD.
Since our method is general, we can easily apply TPLD to the existing self-training based approaches. 
We show consistent improvements over the strong baselines. 
Finally, we achieve new state-of-the-art performances on two standard UDA benchmarks.

\noindent We summarize our contributions as follows:
\begin{enumerate}
\setlength\itemsep{0.3em}
    \item To our best knowledge, it is the first time that pseudo label densification is formally defined and explored in the self-training based domain adaptation.
    \item We present a novel two-phase pseudo label densification framework, called \textbf{TPLD}.
    In particular, for the first phase, we introduce voting-based densification method.
    For the second phase, we propose an easy-hard classification-based densification method. 
    Both phases are complementary in constructing an accurate self-training model.
    \item We propose a new objective function to ease the training. Specifically, we re-formulate the original self-training loss function by incorporating the bootstrapping mechanism.
    \item We conduct extensive ablation studies to thoroughly investigate the impact of our proposals. 
    We apply TPLD to the various existing self-training approaches and achieve new state-of-the-art results on two standard UDA benchmarks.
\end{enumerate}

\section{Related works}
\subsubsection{Domain Adaptation} is a classic problem in computer vision and machine learning. 
It aims to alleviate the performance drop caused by the distribution mismatch in cross-domains.
It is mostly investigated in image classification problems by both conventional methods~\cite{kulis2011you,gopalan2011domain,gong2012geodesic,fernando2013unsupervised,li2017domain} and deep CNN-based methods~\cite{long2015learning,ganin2014unsupervised,ghifary2016deep,sener2016learning,panareda2017open,motiian2017unified,li2017deeper}.
Besides image recognition, domain adaptation is recently being applied other vision tasks such as object detection~\cite{chen2018domain}, depth estimation~\cite{atapour2018real}, and semantic segmentation~\cite{pmlr-v80-hoffman18a}.
In this work, we are particularly interested in \textit{unsupervised} domain adaptation for the task of semantic segmentation.
The primary approach is to minimize the discrepancy between source and target feature distribution using adversarial learning.
This type of approaches is studied on three different levels in practice:
input-level alignment~\cite{learnSi,imagetrans, Chen_2019_CVPR, pmlr-v80-hoffman18a}, intermediate feature-level alignment~\cite{trasdeepadat, fcnwild, vu2019advent, Hong_2018_CVPR, Yawei2019Taking}, and output-level alignment~\cite{tsai2018learning}.
Although these methods are proven to be effective, the potentially meaningful training signals from the target domain are under-utilized.
Therefore, self-training based UDA approaches~\cite{zou2018unsupervised,Zou_2019_ICCV}, described next, emereged recently and came to dominate the performance quickly.
\subsubsection{Self-training} has been initially explored in semi-supervised method~\cite{semitutorial, semientropy}.
Recently, two seminar works~\cite{zou2018unsupervised,Zou_2019_ICCV} have been presented for UDA semantic segmentation.
Unlike adversarial learning approaches, these methods explicitly explore the supervision signals from the target domain. 
The key idea is to use the prediction from the source-trained model as pseudo-labels for the unlabeled data and re-trains the current model in the target domain.
CBST~\cite{zou2018unsupervised} extends this basic idea with class balancing strategy and spatial priors.
CRST~\cite{Zou_2019_ICCV} further adds regularization term in the loss function to prevent overconfident predictions.
In this paper, we also investigate the self-training framework. 
However, different from the previous studies, we see that the spare pseudo label problem is a fundamental limitation of self-training. 
We empirically found that these sparse pseudo-labels inhibit effective learning; thus, the model significantly deviates from the optimal.
We, therefore, propose to densify the sparse pseudo-labels in a two-step gradually. Also, we present a new loss function to handle noisy pseudo labels and reduce optimization difficulties during training. We empirically confirm that our proposals greatly improve the strong state-of-the-art baselines with healthy margins.

\section{Preliminaries}
\subsection{Problem Setting}
Following the common UDA setting, we have full access to the data and labels, $(\boldsymbol{\mathrm{x_{s}}}, \boldsymbol{\mathrm{y_{s}}})$, in the labeled source domain. In contrast, in the unlabeled target domain, we can only utilize the data, $\boldsymbol{\mathrm{x_{t}}}$.
In self-training, we thus train the network to infer pseudo target label, $\boldsymbol{\mathrm{\hat{y}_{t}}}$ = $({\hat{y}_{t}}^{(1)}, ..., {\hat{y}_{t}}^{(K)})$, where $K$ denotes the total number of classes.

\subsection{Self-training for UDA}

\sloppy 
We first revisit the general self-training loss function~\cite{Zou_2019_ICCV} below:
\begin{equation}
    \begin{split}
    \min_\mathrm{\textbf{w}, \hat{\textbf{Y}}_{\textbf{T}}}&\mathcal{L}_{st}\mathrm(\textbf{w}, \hat{\textbf{Y}}_{\textbf{T}})
    = - \sum_{s \in S}\sum_{k=1}^K y_{s}^{(k)}\log p(k| \mathrm{\textbf{x}_{\textbf{s}};\textbf{w}}) \\
    & - \sum_{t \in T}[\sum_{k=1}^K \hat{y}_{t}^{(k)}\log \frac{p(k| \mathrm{\textbf{x}_{\textbf{t}};\textbf{w}})}{\lambda_{k}} - \alpha r_{c}\mathrm(\textbf{w}, \hat{\textbf{Y}}_{\textbf{T}})]\\
    &s.t. \hspace{0.5em}  \hat{y}^{t} \in \Delta^{K-1} \cup \{\textbf{0}\}, \forall{t} \\
    \end{split}
    \label{eq:crst}
\end{equation}




$\boldsymbol{\mathrm{x_{s}}}$ denotes an image in source domain indexed by $s = 1, 2, ...,S$, and 
$\boldsymbol{\mathrm{x_{t}}}$ is an image in target domain indexed by $t = 1, 2, ...,T$. 
$y_{s}^{(k)}$ is ground truth source label for class $k$, and
$\hat{y}_{t}^{(k)}$ is generated pseudo target label.
Note that feasible set of pseudo-label is the union of $\{\textbf{0}\}$ and a probability simplex $\Delta^{K-1}$ (i.e., continuous).
$\boldsymbol{\mathrm{w}}$ is the network weights, and $p(k|\boldsymbol{\mathrm{x}};\boldsymbol{\mathrm{w}})$ indicates the classifier's softmax probability for class $k$. $\lambda_{k}$ is a parameter, controlling pseudo-label selection~\cite{zou2018unsupervised}. 
$\sum_{t \in T} r_{c}\mathrm(\textbf{w}, \hat{\textbf{Y}}_{\textbf{T}})$ is the confidence regularizer and $\alpha \geq 0$ is the weight coefficient. 

We can better understand the equation \eqref{eq:crst} by dividing it into three terms;
The first term is model training on source domain with source labels, $y_{s}$.
The second term is model re-training on target domain with generated target pseudo labels, $\hat{y}_{t}$. 
The last term is confidence regularization, $\alpha r_{c}\mathrm(\textbf{w}, \hat{\textbf{Y}}_{\textbf{T}})$, which prevents over-confident predictions of target pseudo-labels.
The first two terms are equivalent to the CBST formula~\cite{zou2018unsupervised}.
With the additional confidence regularization term, we come up with the CRST formula~\cite{Zou_2019_ICCV}.
In general, there are two types of regularization: label-regularization(e.g, LRENT) and model regularization(e.g, MRKLD).

To minimize Eq.~\eqref{eq:crst}, the optimization algorithm alternatively takes block coordinate descent on both 1) pseudo-label generation and 2) network retraining.
For solving step 1), there is an optimizer formulated as:
\begin{equation}
   \hat{y}^{(k)*}_{t}=\begin{cases}
    1, & \text{if $k = \argmax\limits_{k}\{\frac{p(k|\boldsymbol{\mathrm{x_{t};w}})}{\lambda_{k}}\}$} \\
       & $and$ \hspace*{0.2cm} p(k|\boldsymbol{\mathrm{x_{t};w}}) > \lambda_{k} \\
    0, & \text{otherwise}.
  \end{cases}
  \label{eq:case1}
\end{equation}
If the prediction is confident, $p(k|\boldsymbol{\mathrm{x_{t};w}}) > \lambda_{k}$, it is selected and labeled as a class $k^{*} =  \argmax\limits_{k}\{\frac{p(k|\boldsymbol{\mathrm{x_{t};w}})}{\lambda_{k}}\}$. 
Otherwise, the less confident predictions are set to zero vector $\boldsymbol{0}$. 
For each class k, we determine $\lambda_{k}$ by the confidence value that is selected from the most confident $p$ portion of class k predictions in the entire target set~\cite{zou2018unsupervised}. To avoid selecting unconfident predictions at the early stage, the hyperparameter $p$ is usually set to a low value (i.e., 0.2), and is gradually increased for each additional round. To solve step 2), we use typical gradient-based methods (e.g., SGD). 
For more details, please refer to the original papers~\cite{zou2018unsupervised,Zou_2019_ICCV}.

We see the current self-training approach simply zeroes out the less confident predictions and in turn generates sparse pseudo labels. 
We argue that this limits the power of model representations and could produce sub-optimal model.
Motivated by our empirical observations, we attempt to densify the sparse pseudo labels gradually, and avoid noisy predictions. In this work, we propose TPLD, which alleviates these fundamental issues successfully. 
We show the TPLD can be applied to any type of existing self-training based frameworks, and can consistently boost the performance significantly.


\subsection{Noisy label handling}
To handle noisy predictions, Reed et.al~\cite{article} proposed bootstrapping loss. It is a weighted sum of the standard cross-entropy loss and the (self) entropy loss. In this work, we apply it to the self-training formula as:
\begin{equation}
    \begin{split}
    \sum_{t \in T}\sum_{k=1}^K [\beta \hat{y}_{t}^{(k)} + (1-\beta)\frac{p(k| \mathrm{\textbf{x}_{\textbf{t}};\textbf{w}})}{\lambda_{k}}]\log \frac{p(k| \mathrm{\textbf{x}_{\textbf{t}};\textbf{w}})}{\lambda_{k}}
    \end{split}
    \label{eq:BT}
\end{equation}
Intuitively, it simultaneously encourages the model to predict the correct (pseudo) target label and have high confidence on its prediction.
\begin{figure*}[t]
    \centering 
    \includegraphics[width=340pt, height = 150pt]{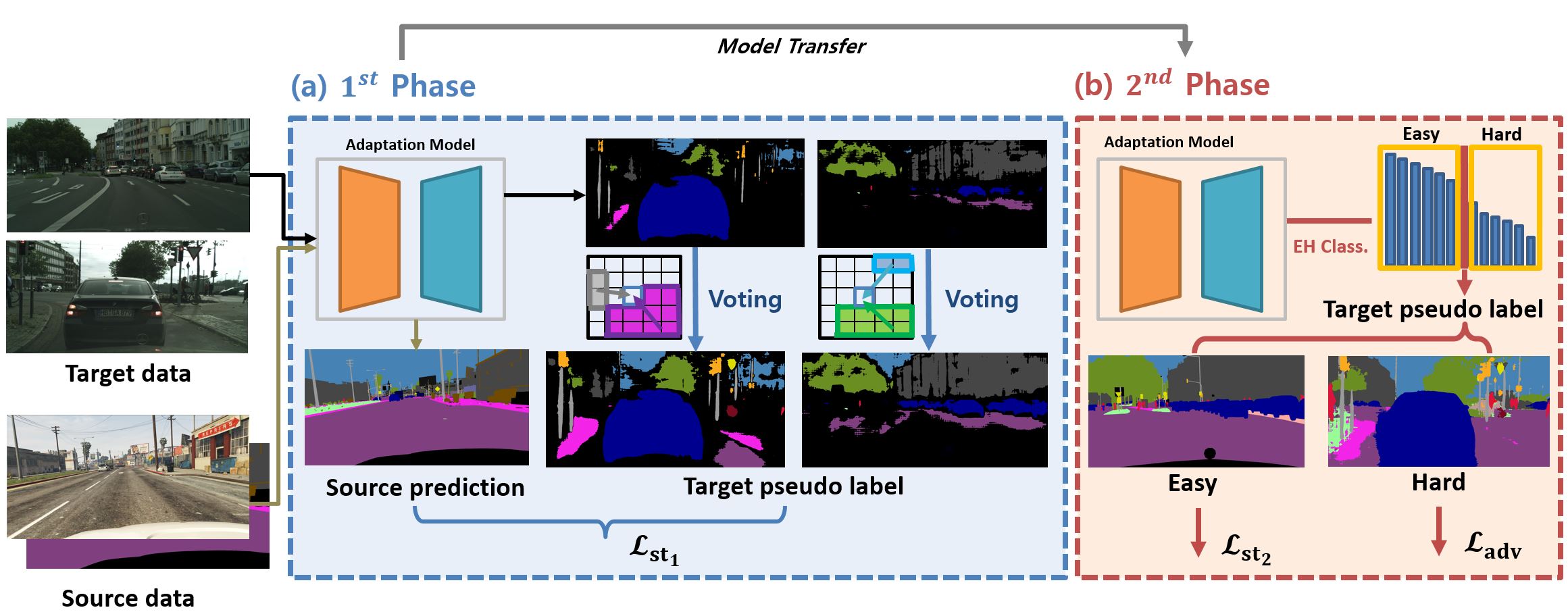}
    \caption{\textbf{The overview of the proposed two-phase pseudo-label densification framework.} (a) \textit{The first phase} utilizes the sliding window based voting in which it propagates neighbor confident predictions to fill in the unlabeled pixels. We use $\mathcal{L}_{st_{1}}$ to train the model in the first phase. (b) \textit{The second phase} employs confidence-based easy-hard classification (EH class.) along with the hard-to-easy adversarial learning. This allows the model to utilize full pseudo labels for easy samples while pushing hard samples to be like easy. We use both $\mathcal{L}_{st_{2}}$ and $\mathcal{L}_{adv}$ to train the model in the second phase.}
    \label{fig:architecture}
\end{figure*}

\section{Method}
The overview of our two-phase pseudo-label densification algorithm is shown in~\figref{fig:architecture}.
For the first phase, we design a sliding window-based voting method to propagate the confident predictions. 
After enough training, we enter the second phase. Here, we present confidence based easy-hard classification and hard/easy adversarial learning. For both phases, we use the proposed bootstrapped self-training loss (Eq.~\eqref{eq:BT}). We detail each phase below.

\subsection{\nth{1} phase: Voting based Densification}
\begin{figure}[t]
    \centering
    \includegraphics[width=250pt, height = 100pt]{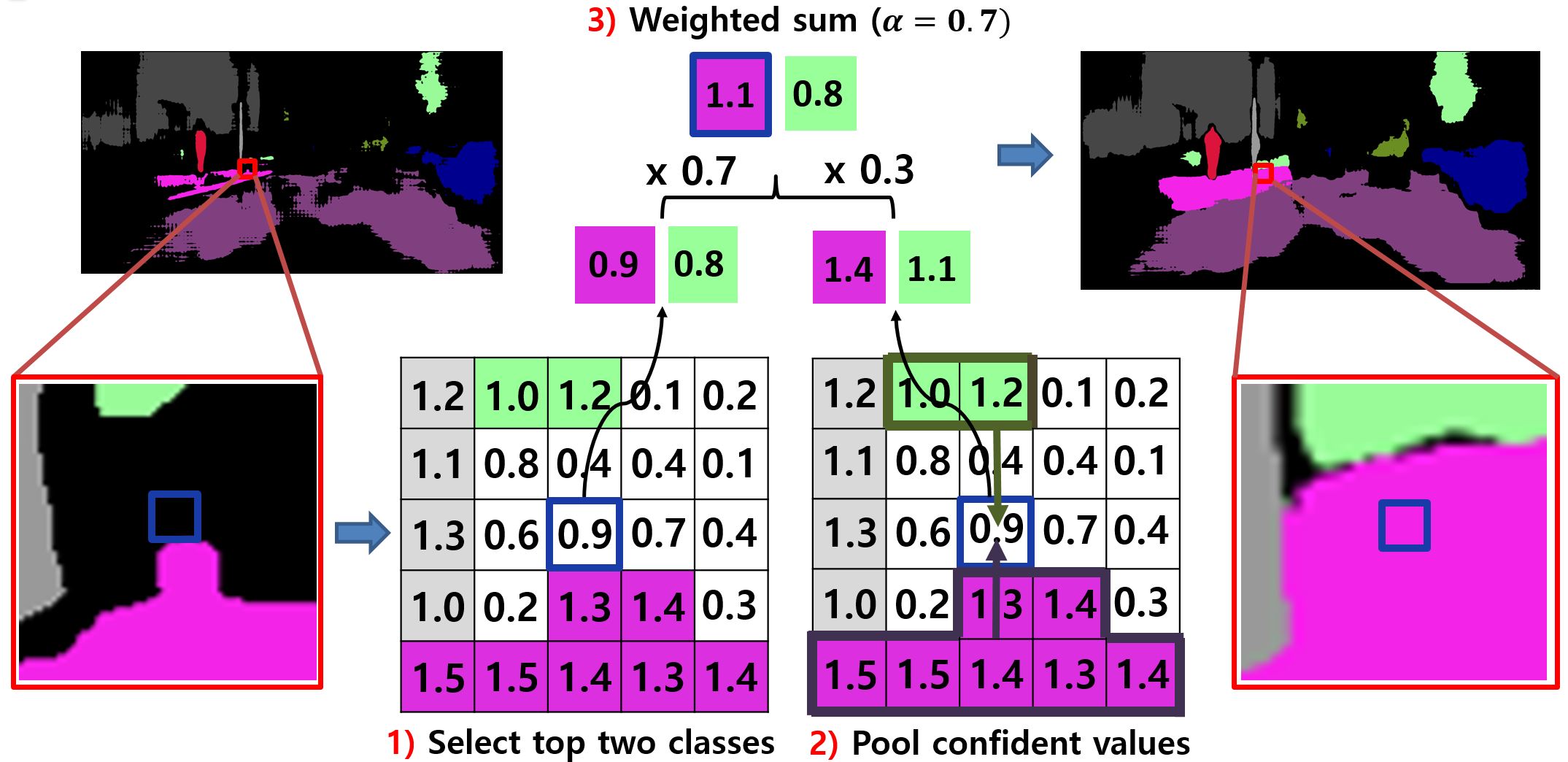}
    \caption{\textbf{The overall procedure of the voting-based densification.} We describe the process in three steps. 1) We find the top two competing classes on the unlabeled pixel, 2) We pool neighboring confident values for these classes, 3) We combine the original prediction values and the pooled values (weighted-sum with hyperparameter $\alpha$). We pick the bigger one and assign the corresponding class if it passes the thresholding criteria. We repeat this process by sliding the window across the images.}
    \label{fig:voting_22}
\end{figure}

As mentioned above, pseudo labels are generated only when the sample's prediction is confident (Eq.~\eqref{eq:case1}).
Specifically, the most confident $p$ portion of predictions are selected class-wise.
Because the hyperparameter $p$ is set to a low value in practice, pseudo labels are inherently sparse during training.
To overcome this issue, we present a sliding window-based voting, in which it relaxes the current hard-thresholding and propagates the confident predictions based on the intrinsic spatial correlations in the image.
We attempt to utilize the fact that neighboring pixels tend to be alike.
To efficiently employ this local spatial regularity in the image, we adopt the sliding-window approach.
We detail the process in~\figref{fig:voting_22}.
Given the window with the unlabeled pixel at the center, we gather the neighboring confident prediction values (voting).
To be more specific, for the unlabeled pixel, we first obtain the top two competing classes (i.e., classes with highest and second-highest prediction values, which would have caused ambiguity in deciding the correct label) (\figref{fig:voting_22}-\textcolor{red}{1}), and then pool the neighboring confident values for these classes (\figref{fig:voting_22}-\textcolor{red}{2}). The spatially-pooled prediction values are then weighted sum with the original prediction values (\figref{fig:voting_22}-\textcolor{red}{3}). Among the two values, we choose the bigger one. Finally, if it is above the threshold, we select the according class as a pseudo label. 
Note that, we use normalized prediction values (i.e., $\frac{p(k|\boldsymbol{\mathrm{x_{t};w}})}{\lambda_{k}}$) during the voting process, thus the thresholding criteria is $\frac{p(k|\boldsymbol{\mathrm{x_{t};w}})}{\lambda_{k}} > 1$.
Otherwise, it continues to be a zero vector. 

\begin{figure*}[t]
    \centering 
    \includegraphics[width=1.0\textwidth]{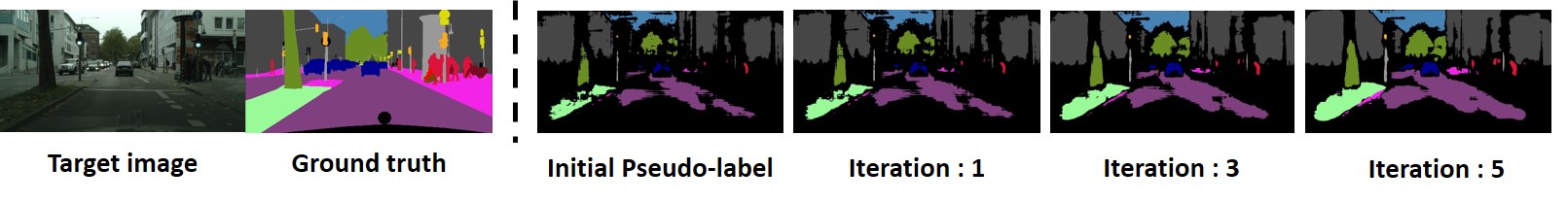}
    \caption{\textbf{Voting based densification results by iteration.} We can see the initial sparse pseudo label becomes dense as iteration number increases. Though it may bring noisy predictions. We set the total iteration number to 3 after conducting parameter analysis in~\tabref{tab:abla}.}
    \label{fig:3}
\end{figure*}

We call the above whole process voting-based densification. We abbreviate it as $\mathrm{\textbf{Voting}}$.
We iterate over total 3 times with the window size of $57\times 57$. Those hyperparameters are set through the parameter analysis (see~\tabref{tab:VF_VN}).
The qualitative voting results are shown in~\figref{fig:3}.
We can clearly see that the initial sparse pseudo label gradually becomes dense.
The pseudo label generation in the 1st phase can be summarized as:
\begin{equation}
    \begin{split}
   \hat{y}^{(k)*}_{t}=\begin{cases}
    1, & \hspace*{-0.2cm} \text{if $k = \argmax\limits_{k}\{\frac{p(k|\boldsymbol{\mathrm{x_{t};w}})}{\lambda_{k}}\}$} \\
       & \hspace*{-0.2cm} $and$ \hspace*{0.2cm} p(k|\boldsymbol{\mathrm{x_{t};w}}) > \lambda_{k} \\
    \mathrm{\textbf{Voting}(\frac{p(k|\boldsymbol{\mathrm{x_{t};w}})}{\lambda_{k}})}, & \hspace*{-0.2cm} \text{otherwise}
  \end{cases}
  \label{eq:eq_4}
  \end{split}
\end{equation}
\subsubsection{Objective function for the \nth{1} phase}
To effectively train the model under the existence of noisy pseudo labels, we introduce bootstrapping (\eqnref{eq:BT}) in our final objective function.
The original self-training objective function can be re-formulated as the following:
\begin{equation}
    \begin{split}
    \min_\mathrm{\textbf{w}, \hat{\textbf{Y}}_{\textbf{T}}}&\mathcal{L}_{st_{1}}\mathrm(\textbf{w}, \hat{\textbf{Y}}_{\textbf{T}})
    = - \sum_{s \in S}\sum_{k=1}^K y_{s}^{(k)}\log p(k| \mathrm{\textbf{x}_{\textbf{s}};\textbf{w}}) \\
    & - \sum_{t \in T}[\sum_{k=1}^K \{\beta \hat{y}_{t}^{(k)} + (1-\beta)\frac{p(k| \mathrm{\textbf{x}_{\textbf{t}};\textbf{w}})}{\lambda_{k}}\} \log \frac{p(k| \mathrm{\textbf{x}_{\textbf{t}};\textbf{w}})}{\lambda_{k}} \\
    & - \alpha r_{c}\mathrm(\textbf{w}, \hat{\textbf{Y}}_{\textbf{T}})]\\
    &s.t. \hspace{0.5em}  \hat{y}^{t} \in \Delta^{K-1} \cup \{\textbf{0}\}, \forall{t} \\
    \end{split}
    \label{eq:eq_5}
\end{equation}

As a result, the target domain training benefits from both densified pseudo label and bootstrapped training.

\subsection{\nth{2} phase: Easy-Hard Classification based Densification}

\begin{figure*}[t]
    \centering 
    \includegraphics[width=1.0\textwidth]{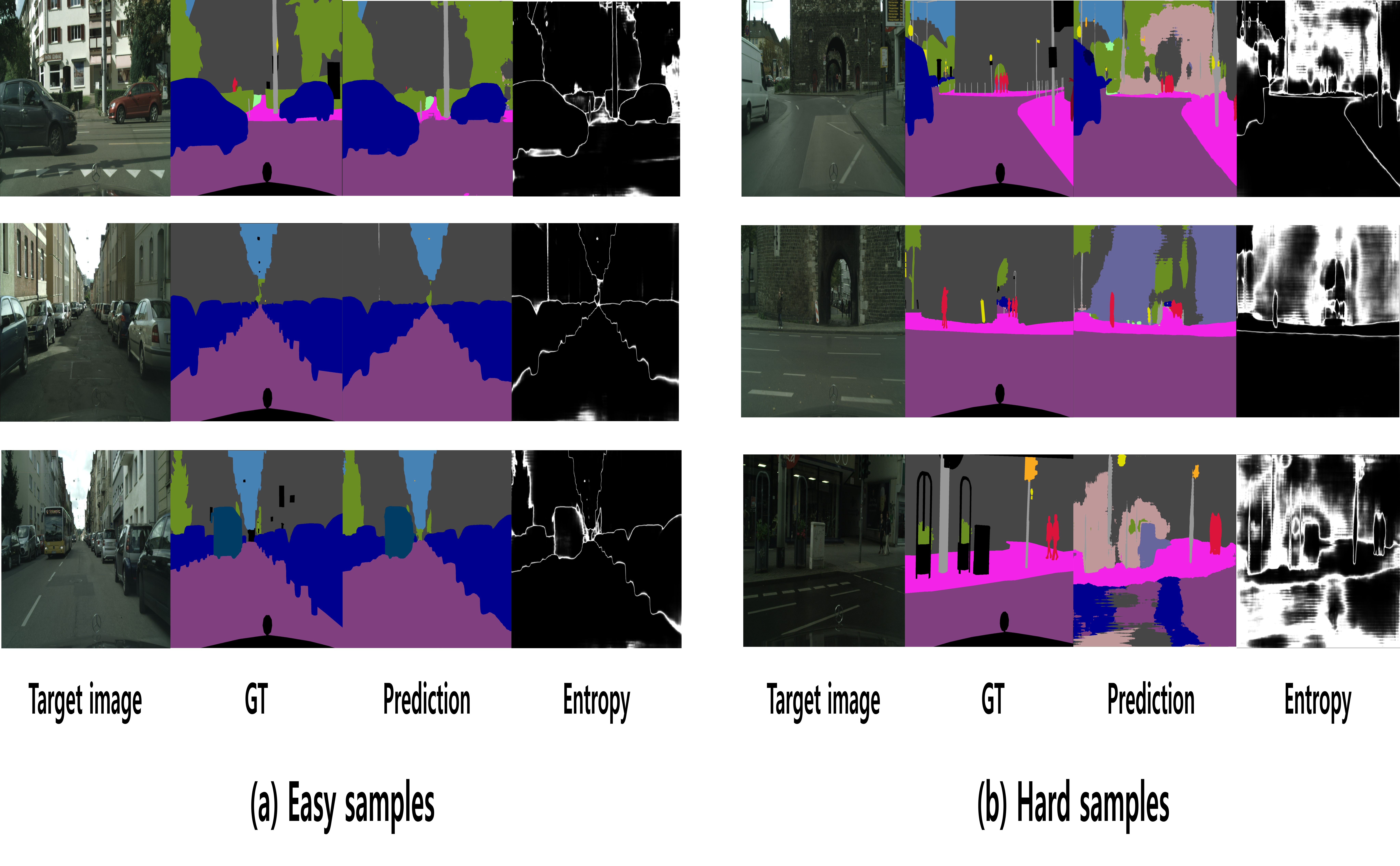}
    \caption{\textbf{Qualitative easy and hard samples.} 
    For the illustration, we randomly selected three samples from each. Note that easy samples are near to the ground truth with low entropy values, whereas hard samples are far from the ground truth and have high entropy values. Therefore, in the second phase, we train easy samples with their full-pseudo labels and make hard samples to be easy using adversarial loss.}
    \label{fig:easyhard}
\end{figure*}

As the predictions of model can be trusted more over time, we now attempt to use full pseudo-labels.
One may attempt to use voting multiple times for full densificaiton.
However, the experimental evidence shown in~\tabref{tab:VF_VN} proves that it is hard for voting to generate fully densified pseudo labels. 
By construction, the voting is operated with a local window, which can only capture and process local predictions. Thus, iterating the voting process multiple times brings some extent of smoothing effect and noisy predictions.
We, therefore, present another phase which enables full-pseudo label training.
Our key idea is to consider the confidence on image-level and classify the images into two groups: easy and hard. For the easy, confident samples, we utilize their full predictions, while for the hard samples, we instead enforce hard-to-easy adaption.
Indeed, we observe that the easy samples are near to the ground truth and vice versa (see~\figref{fig:easyhard}).

To reasonably categorize target samples into easy and hard, we present effective criteria.
For a particular image $\textbf{t}$, we define a confidence score as
    ${conf}_{\mathrm{t}} = \frac{1}{K'}\sum_{k=1}^{K'} \frac{N_{\mathrm{t}}^{k*}}{N_{\mathrm{t}}^{k}} \cdot \frac{1}{\lambda_{k}},$
where $N_{\mathrm{t}}^{k}$ is the total number of pixels predicted as class $k$.
Among $N_{\mathrm{t}}^{k}$, we count the number of pixels that have higher prediction values than the class-wise thresholding value $\lambda_{k}$~\cite{zou2018unsupervised}, and is set to $N_{\mathrm{t}}^{k*}$.
As a result, the ratio $\frac{N_{\mathrm{t}}^{k*}}{N_{\mathrm{t}}^{k}}$ indicates how well the model predicts confident values for each class $k$. We average these values with $K'$, which is the total number of (predicted) confident classes.
Thus, the higher the value, we can say that the model is more confident with that target image (i.e., easy).
Note that, we multiply $\frac{1}{\lambda_{k}}$ to avoid sampling too easy images and instead encourage sampling of images with rare classes. We compute these confidence scores for every target image.
In practice, we picked up the top $q$ portion as easy samples and consider the rest as hard samples for the training.
We initially set $q$ to 30\% and add 5\% in each round.

\subsubsection{Objective function for the \nth{2} phase}
After classifying target images into easy and hard samples, we apply different objective functions to each.
For the easy samples, we utilize full pseudo label predictions and employ bootstrapping loss for training (Eq.~\ref{eq:BT}).
For the hard samples, we instead adopt adversarial learning to push hard examples to be like easy samples (i.e., feature alignment). We describe the details below.
\subsubsection{Easy sample training} To effectively generate full pseudo labels, we calibrate the prediction values.
Specifically, the full pseudo-label generation of easy samples is formulated as:
\begin{equation}
   \hat{y}^{(k)*}_{t_{e}}=\begin{cases}
    1, & \text{if $k = \argmax\limits_{k}\{\frac{p(k\mid x_{t};w)}{\lambda_{k}}\}$} \\
       & $and$ \hspace*{0.2cm} p(k|\boldsymbol{x_{t};w}) > \lambda_{k} \\
    \bigl(\frac{p(k|\boldsymbol{x_{t};w})}{\lambda_{k}}\bigr)^{\gamma}, & \text{otherwise}.
  \end{cases}
  \label{eq:eq_7}
\end{equation}
Note that the prediction value is calibrated with the hyper parameter $\gamma$, which is set to 2 empirically (see~\tabref{tab:gamma}).
We then train the model using the following bootstrapping loss:
\begin{equation}
    \begin{split}
    &\min_\mathrm{\textbf{w}, \hat{\textbf{Y}}_{\textbf{T}}}\mathcal{L}_{st_{2}}\mathrm(\textbf{w}, \hat{\textbf{Y}}_{\textbf{T}}) \\
    &= - \sum_{t \in T}[\sum_{k=1}^K \{\beta \hat{y}_{t}^{(k)} + (1-\beta)\frac{p(k| \mathrm{\textbf{x}_{\textbf{t}};\textbf{w}})}{\lambda_{k}}\} \log \frac{p(k| \mathrm{\textbf{x}_{\textbf{t}};\textbf{w}})}{\lambda_{k}} \\
    &s.t. \hspace{0.5em}  \hat{y}^{t} \in \Delta^{K-1} \cup \{\textbf{0}\}, \forall{t} \\
    \end{split}
    \label{eq:eq_8}
\end{equation}
\subsubsection{Hard sample training} To minimize the gap between easy ($e$) and hard ($h$) samples in the target domain, we propose intra-domain adversarial loss, $\mathcal{L}_{adv}$. In order to align the feature from hard to easy, the discriminator $D_{intra}$ is trained to discriminate that the target weighted self-information map $I_{t}$~\cite{vu2019advent} is whether from easy samples or hard samples.
The learning objective of the discriminator is:
\begin{equation}
    \min_{\theta_{D_{intra}}}\frac{1}{\left|e\right|}\sum_{e}L_{D_{intra}}(I_{e}, 1) + \frac{1}{\left|h\right|}\sum_{h}L_{D_{intra}}(I_{h}, 0)
    \label{eq:eq_9}
\end{equation}
and the adversarial objective to train the segmentation network is:
\begin{equation}
    \min_{\theta_{seg}}\frac{1}{\left|h\right|}\sum_{h}L_{D_{intra}}(I_{h}, 1)
    \label{eq:eq_10}
\end{equation}

\section{Experiments}

\subsection{Dataset}
We evaluate our model on the most common adaptation benchmarks. 
1) GTA5~\cite{Richter_2016_ECCV} to Cityscapes~\cite{Cordts2016Cityscapes} and 2) SYNTHIA~\cite{ros2016synthia} to Cityscapes. 
GTA5 and SYNTHIA contain 24966 and 9,400 synthetic images, respectively. 
Following the standard protocols, we adapt the model to the Cityscapes training set and evaluate the performance on the validation set.

\subsection{Implementation details}
To push the state-of-the-art benchmark performances, we apply TPLD to the CRST-MRKLD framework~\cite{Zou_2019_ICCV}.
For the backbones, we use VGG-16~\cite{vgg} and ResNet-101~\cite{resnet}. 
For the segmentation models, we adopt different versions of deeplab; deeplab-v2~\cite{deep2} and deeplab-v3~\cite{deep3}.
We pretrain the model on ImageNet~\cite{imagenet} and fine-tune on source domain images using SGD. 
We train the model total 9 rounds: 6 rounds for the first phase training and 3 rounds for the second phase training.
The detailed training settings are the followings: 
For the source domain pre-training, we use learning rate of $2.5\times10^{-4}$, weight decay of $5\times10^{-4}$, momentum of 0.9, batch size of 2, patch size of $512\times 1024$, multiscale training augmentation (0.5 - 1.5), and horizontal flipping. 
For the self-training, we adopt SGD with the learning rate of $5\times10^{-5}$.

\subsection{Main Results}

\subsubsection{GTA5 $\to$ Cityscapes:} 
\begin{figure*}[t]
    \centering 
    \includegraphics[width=0.999\textwidth]{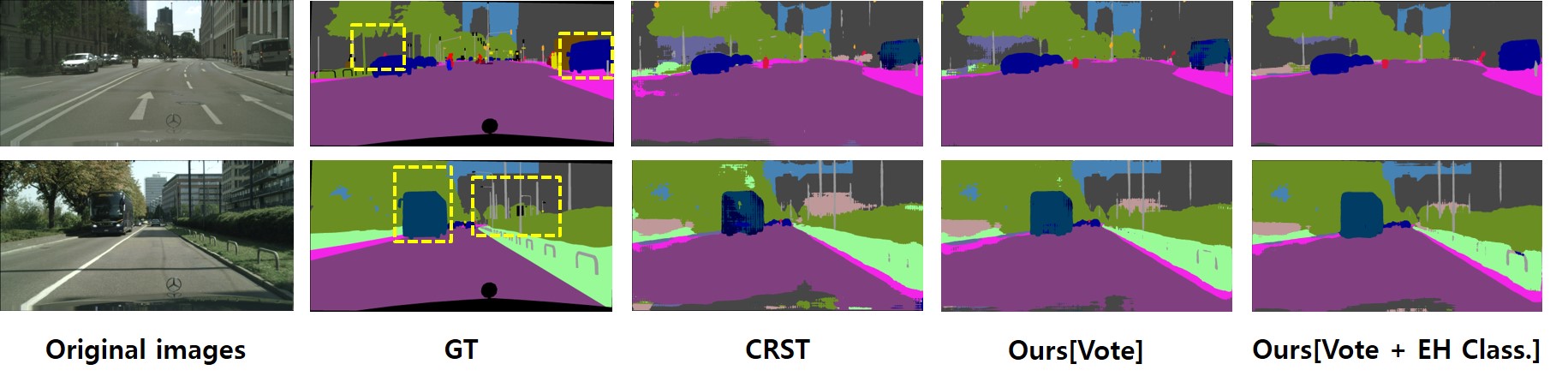}
    \caption{\textbf{Qualitative results on GTA5 $\to$ Cityscapes.} We can clearly see that our full model generates the most visually pleasable results.}
    \label{fig:quality}
\end{figure*}

Table~\ref{table:gta5} summarizes the adaptation performance of TPLD and other state-of-the-art methods~\cite{tsai2018learning,Yawei2019Taking,vu2019advent, zou2018unsupervised,Zou_2019_ICCV}.
We can obviously see that TPLD outperforms state-of-the-art approaches in all cases.
For example, with Deeplab-v2 and ResNet-101 backbone, our TPLD significantly outperforms CRST by 4.2\%. 
Moreover, to analyze the effect on rare classes, we also put rare-class mIoU.
With the R-mIoU metric, we see the improvement is even much higher; 4.8\%.
We provide qualitative results in Figure~\ref{fig:quality}. 
Clearly, our final model generates the most visually pleasurable results.

\subsubsection{SYNTHIA $\to$ Cityscapes:} Table~\ref{table:syn} shows the adaptation results with SYNTHIA. 
Our approach again achieves the best performance among all the other methods. 
Specifically, with Deeplab-v3 and ResNet101 backbone, we greatly improve the baseline performance of 48.1\% mIoU to 55.7\%  mIoU.

\begin{table*}[t]
\renewcommand{\arraystretch}{1.2}
\begin{center}
\centering
\resizebox{\textwidth}{!}{
\begin{tabular}{l l c c c c c c c c c c c c c c c c c c c|c|c}
\toprule
\multicolumn{23}{c}{GTA5 $\to$ Cityscapes}\\
\midrule
Method & Seg Model & Road & SW & Build & \textcolor{mypink3}{Wall} & \textcolor{mypink3}{Fence} & \textcolor{mypink3}{Pole} & \textcolor{mypink3}{TL} & \textcolor{mypink3}{TS} & Veg. & \textcolor{mypink3}{Terrain} & Sky & PR & \textcolor{mypink3}{Rider} & Car & \textcolor{mypink3}{Truck} & \textcolor{mypink3}{Bus} & \textcolor{mypink3}{Train} & \textcolor{mypink3}{Motor} & \textcolor{mypink3}{Bike} & mIoU & R-mIoU \\
\midrule
Source & \multirow{4}{*}{Deeplabv2-V}  & 52.6 & 20.7 & 56.0 & 6.0 & 9.8 & 22.9 & 8.1 & 1.4 & 77.2 & 11.0 & \textbf{35.0} & 41.5 & 2.7 & 52.1 & 2.1 & 0.0 & 0.0 & 4.7 & 0.3 & 21.3 & 5.8 \\
CBST~\cite{zou2018unsupervised} &  & \textbf{84.2} & 41.4 & 71.9 & 15.5 & \textbf{18.1} & \textbf{30.8} & 25.4 & 9.2 & 77.6 & 15.2 & 29.6 & \textbf{49.3} & 6.0 & 78.0 & \textbf{4.0} & 4.5 & 0.3 & 10.4 & 11.6 & 30.7 & 12.6 \\
CRST(MRKLD)~\cite{Zou_2019_ICCV} &  & 81.7 & 46.1 & 70.2 & 10.7 & 11.2 & 30.4 & 26.9 & \textbf{15.8} & 75.4 & 18.3 & 24.8 & 48.6 & 10.9 & 77.8 & 2.9 & 13.3 & 1.1 & 10.7 & 31.4 & 32.0 & 15.3 \\
\rowcolor{lightgray} CRST(MRKLD) + TPLD &   & 83.5 & \textbf{49.9} & \textbf{72.3} & \textbf{17.6} & 10.7 & 29.6 & \textbf{28.3} & 9.0 & \textbf{78.2} & \textbf{20.1} & 25.7 & 47.4 & \textbf{13.3} & \textbf{79.6} & 3.3 & \textbf{19.3} & \textbf{1.3} & \textbf{14.3} & \textbf{33.5} & \textbf{34.1} & \textbf{16.7}\\
\midrule\midrule
Adapt-SegMap~\cite{tsai2018learning} & \multirow{3}{*}{Deeplabv2-R}  & 86.5 & 36.0 & 79.9 & 23.4 & 23.3 & 35.2 & 14.8 & 14.8 & 83.4 & 33.3 & 75.6 & 58.5 & 27.6 & 73.7 & 32.5 & 35.4 & 3.9 & 30.1 & 28.1 & 42.4 & 25.2 \\
CLAN~\cite{Yawei2019Taking} &   & 87.0 & 27.1 & 79.6 & 27.3 & 23.3 & 28.3 & \textbf{35.5} & 24.2 & 83.6 & 27.4
 & 74.2 & 58.6 & 28.0 & 76.2 & 33.1 & 36.7 & 6.7 & 31.9 & 31.4 & 43.2 & 27.8 \\
ADVENT~\cite{vu2019advent} &  & 89.9 & 36.5 & 81.2 & 29.2 & \textbf{25.2} & 28.5 & 32.3 & 22.4 & 83.9 & 34.0 & 77.1 & 57.4 & 27.9 & 83.7 & 29.4 & 39.1 & 1.5 & 28.4 & 23.3 & 43.8 & 26.8 \\
\midrule
Source & \multirow{4}{*}{Deeplabv2-R} & 71.3 & 19.2 & 69.1 & 18.4 & 10.0 & 35.7 & 27.3 & 6.8 & 79.6 & 24.8 & 72.1 & 57.6 & 19.5 & 55.5 & 15.5 & 15.1 & 11.7 & 21.1 & 12.0 & 33.3 & 18.2\\
CBST~\cite{zou2018unsupervised}& & 91.8 & 53.5 & 80.5 & 32.7 & 21.0 & 34.0 & 28.9 & 20.4 & 83.9 & 34.2 & 80.9 & 53.1 & 24.0 & 82.7 & 30.3 & 35.9 & 16.0 & 25.9 & 42.8 & 45.9 & 28.9  \\
CRST(MRKLD)~\cite{Zou_2019_ICCV} & & 91.3 & 56.1 & 79.8 & 30.6 & 18.9 & 39.0 & 35.1 & 24.0 & 84.2 & 30.0 & 74.0 & \textbf{62.1} & \textbf{28.2} & 82.6 & 23.6 & 31.8 & 24.2 & 32.2 & 46.3 & 47.0 & 30.3\\
\rowcolor{lightgray} CRST(MRKLD) + TPLD & & \textbf{94.2} & \textbf{60.5} & \textbf{82.8} & \textbf{36.6} & 16.6 & \textbf{39.3} & 29.0 & \textbf{25.5} & \textbf{85.6} & \textbf{44.9} & \textbf{84.4} & 60.6 & 27.4 & \textbf{84.1} & \textbf{37.0} & \textbf{47.0} & \textbf{31.2} & \textbf{36.1} & \textbf{50.3} & \textbf{51.2} & \textbf{35.1}  \\
\midrule\midrule

Source & \multirow{4}{*}{Deeplabv3-R}  & 80.3 & 17.6 & 75.8 & 18.0 & 24.5 & 19.7 & 34.9 & 19.0 & 83.2 & 15.8 & 63.7 & 57.2 & 22.8 & 73.4 & 36.6 & 21.0 & 0.0 & 19.0 & \textbf{0.1} & 35.9 & 19.3\\
CBST~\cite{zou2018unsupervised}&    & \textbf{86.9} & 33.9 & \textbf{80.0} & 28.8 & \textbf{26.2} & 30.2 & 36.9 & 20.4 & 84.6 & 16.3 & 72.1 & 53.3 & 19.8 & 82.8 & 34.1 & 43.8 & 0.0 & 13.0 & 0.0 & 40.2 & 22.5 \\
CRST(MRKLD)~\cite{Zou_2019_ICCV} &   & 85.9 & 40.4 & 76.9 & 27.5 & 21.6 & \textbf{35.0} & \textbf{39.0} & \textbf{25.6} & 84.0 & 20.2 & 71.8 & 55.3 & 23.2 & 83.2 & \textbf{38.8} & 43.2 & 0.0 & 10.3 & 0.0 & 41.2 & 23.7 \\
\rowcolor{lightgray} CRST(MRKLD) + TPLD &    & 83.2 & \textbf{46.3} & 74.9 & \textbf{29.8} & 21.3 & 33.1 & 36.0 & 24.2 & \textbf{86.7} & \textbf{43.2} & \textbf{87.1} & \textbf{58.7} & \textbf{24.0} & \textbf{84.0} & 36.9 & \textbf{49.7} & 0.0 & \textbf{29.7} & 0.0 & \textbf{44.7} & \textbf{27.3} \\
\bottomrule
\end{tabular}
}
\end{center}
\caption{\textbf{Experimental results on GTA5 $\rightarrow$ Cityscapes.} "V" and "R" denote VGG-16 and ResNet-101 respectively. We highlight the rare classes~\cite{Yawei2019Taking} and compute Rare class mIoU (R-mIoU) as well.}
\label{table:gta5}
\end{table*}
\begin{table*}[t]
\renewcommand{\arraystretch}{1.2}
\begin{center}

\resizebox{\textwidth}{!}{
\begin{tabular}{l l c c c c c c c c c c c c c c c c |c c |c}
\toprule
\multicolumn{21}{c}{SYNTHIA $\to$ Cityscapes}\\
\midrule
Method & Seg Model  & Road & SW & Build & \textcolor{mypink3}{Wall}$\ast$ & \textcolor{mypink3}{Fence} & \textcolor{mypink3}{Pole}$\ast$ & \textcolor{mypink3}{TL} & \textcolor{mypink3}{TS} & Veg. & Sky & PR & \textcolor{mypink3}{Rider} & Car & \textcolor{mypink3}{Bus} & \textcolor{mypink3}{Motor} & \textcolor{mypink3}{Bike} & mIoU & mIoU$\ast$ & R-mIoU\\
\midrule
Source & \multirow{4}{*}{Deeplabv2-V} & 41.5 & 16.6 & 38.3 & 0.2 & 0.0 & 22.6 & 0.1 & 4.9 & 66.5 & 64.7 & 44.9 & 1.7 & 60.7 & 3.3 & 0.0 & 0.6 & 22.9 & 26.4 & 4.3\\
CBST~\cite{zou2018unsupervised} &   & 75.7 & 32.3 & 70.2 & 3.5 & 0.0 & 28.6 & 1.4 & 9.0 & 79.8 & 65.6 & 52.9 & 13.7 & 65.8 & \textbf{9.1} & \textbf{1.5} & 36.4 & 34.1 & 39.5 & 11.5\\
CRST(MRKLD)~\cite{Zou_2019_ICCV} &   & 75.1 & 33.5 & 70.8 & 5.6 & 0.0 & 28.7 & \textbf{2.0} & \textbf{9.7} & 78.9 & \textbf{72.5} & 51.7 & 11.6 & 63.4 & 7.3 & 1.4 & 38.6 & 34.4 & 39.7 & 11.7\\
\rowcolor{lightgray} CRST(MRKLD) + TPLD &  & \textbf{81.3} & \textbf{34.5} & \textbf{73.3} & 11.9 & 0.0 & 26.9 & 0.2 & 6.3 & \textbf{79.9} & 71.2 & \textbf{55.1} & \textbf{14.2} & \textbf{73.6} & 5.7 & 0.5 & \textbf{41.7}& \textbf{36.0} & \textbf{41.3} & \textbf{11.9}\\
\midrule\midrule
Adapt-SegMap~\cite{tsai2018learning} & \multirow{3}{*}{Deeplabv2-R}   & 84.3 & 42.7 & 77.5 & - & - & - & 4.7 & 7.0 & 77.9 & 82.5 & 54.3 & 21.0 & 72.3 & 32.2 & 18.9 & 32.3 & - & 46.7 & -   \\
ADVENT~\cite{vu2019advent} &   & \textbf{87.0} & 44.1 & 79.7 & 9.6 & 0.6 & 24.3 & 4.8 & 7.2 & 80.1 & \textbf{83.6} & 56.4 & 23.7 & 72.7 & 32.6 & 12.8 & 33.7 & 40.8 & 47.6 & 16.6    \\
CLAN~\cite{Yawei2019Taking} &  & 81.3 & 37.3 & 80.1 & - & - & - & 16.1 & 13.7 & 78.2 & 81.5 & 53.4 & 21.2 & 73.0 & 32.9 & 22.6 & 30.7 & - & 47.8 & - \\
\midrule
Source & \multirow{4}{*}{Deeplabv2-R} & 45.9 & 21.4 & 63.0 & 7.3 & 0.0 & 33.6 & 4.5 & 14.4 & \textbf{81.6} & 79.7 & 55.3 & 16.7 & 67.5 & 21.3 & 7.5 & 19.0 & 33.7 & 38.3 & 13.8  \\ 
CBST~\cite{zou2018unsupervised}&   & 68.0 & 29.9 & 76.3 & 10.8 & 1.4 & 33.9 & \textbf{22.8} & 29.5 & 77.6 & 78.3 & 60.6 & 28.3 & 81.6 & 23.5 & 18.8 & 39.8 & 42.6 & 48.9 & 23.2  \\
CRST(MRKLD)~\cite{Zou_2019_ICCV} &   & 67.7 & 32.2 & 73.9 & 10.7 & 1.6 & 37.4 & 22.2 & \textbf{31.2} & 80.8 & 80.5 & \textbf{60.8} & \textbf{29.1} & 82.8 & 25.0 & 19.4 & \textbf{45.3} & 43.8 & 50.1 & 24.7\\
\rowcolor{lightgray} CRST(MRKLD) + TPLD &  & 80.9 & \textbf{44.3} & \textbf{82.2} & 19.9 & 0.3 & 40.6 & 20.5 & 30.1 & 77.2 & 80.9 & 60.6 & 25.5 & \textbf{84.8} & \textbf{41.1} & \textbf{24.7} & 43.7 & \textbf{47.3} & \textbf{53.5} & \textbf{27.4}\\
\midrule\midrule
Source & \multirow{4}{*}{Deeplabv3-R} & 45.5 & 19.0 & 71.3 & 6.2 & 0.0 & 27.4 & 11.3 & 15.3 & 79.4 & 79.4 & 58.3 & 9.2 & 79.7 & 33.0 & 6.0 & 8.8 & 34.4 & 39.7 & 13.0  \\ 
CBST~\cite{zou2018unsupervised}&   & 45.2 & 19.4 & \textbf{81.8} & 15.7 & 0.2 & 33.3 & \textbf{20.8} & 24.9 & 85.0 & 82.2 & 64.6 & 26.7 & \textbf{84.8} & 48.8 & 22.9 & 43.9 & 43.8 & 50.1 & 26.4  \\
CRST(MRKLD)~\cite{Zou_2019_ICCV} &  & 52.3 & 21.9 & 80.0 & 17.2 & 0.8 & 32.4 & 17.9 & \textbf{31.1} & 84.8 & 83.5 & 63.7 & 28.5 & 83.1 & 37.2 & 19.1 & 52.5 & 44.1 & 50.4 & 26.3\\
\rowcolor{lightgray} CRST(MRKLD) + TPLD &  & \textbf{70.9} & \textbf{29.5} & 80.6 & 18.4 & 0.4 & 26.6 & 19.9 & 30.9 & \textbf{85.5} & \textbf{86.3} & \textbf{66.0} & \textbf{32.9} & 84.4 & \textbf{51.1} & \textbf{29.3} & \textbf{56.2} & \textbf{48.1} & \textbf{55.7} & \textbf{29.5}\\
\bottomrule
\end{tabular}
}
\end{center}
\caption{\textbf{Experimental results on SYNTHIA $\rightarrow$ Cityscapes.} mIoU$\ast$ is computed with 13 classes out of total 16 classes except the classes with$\ast$.}
\label{table:syn}
\end{table*}

\subsubsection{Combining with existing self-training methods}
We see the proposed TPLD is general, thus can be easily applied to the existing self-training based methods.
In this experiment, we combine the TPLD with three different self-training approaches: CBST~\cite{zou2018unsupervised}, CRST with label regularization (LRENT)~\cite{Zou_2019_ICCV}, and CRST with model regularization (MRKLD)~\cite{Zou_2019_ICCV}.
The results are summarized in~\tabref{tab:self_apply}.
We observe that TPLD consistently improves the performance of all the baselines.
The positive results imply that the sparse pseudo-label is indeed a fundamental problem in self-training, and the previous works notably overlooked this problem. We show that the proposed concept of two-phase pseudo-label densificaiton effectively addresses the issue.

\begin{table*}[t]
 \centering
  \setlength{\tabcolsep}{5pt}
  \subfloat[\scriptsize GTA5 $\rightarrow$ Cityscapes]{%
  {\renewcommand\arraystretch{1.1}
    \resizebox{0.45\textwidth}{!}{%
        \begin{tabular}{c c c c}
            \toprule
            \multicolumn{4}{c}{GTA5 $\rightarrow$ Cityscapes (19 categories)} \\
            \midrule
            Method & Base   & + TPLD      &$\triangle$\\                                         
            \midrule
            CBST~\cite{zou2018unsupervised}  &45.9&47.8&+1.9\\
            CRST(LRENT)~\cite{Zou_2019_ICCV} &45.9&47.3&+1.4 \\
            CRST(MRKLD)~\cite{Zou_2019_ICCV} &47.0&51.2&+4.2 \\
            \bottomrule
        \end{tabular}
        }
  }
 }
\setlength{\tabcolsep}{5pt}
  \subfloat[\scriptsize SYNTHIA $\rightarrow$ Cityscapes ]{%
  {\renewcommand\arraystretch{1.1}
    \resizebox{0.45\textwidth}{!}{%
        \begin{tabular}{c c c c}
            \toprule
            \multicolumn{4}{c}{SYNTHIA $\rightarrow$ Cityscapes (16 categories)} \\
            \midrule
            Method & Base   & + TPLD      &$\triangle$\\                                         
            \midrule
            CBST~\cite{zou2018unsupervised}  &42.6&45.6& +3.0\\
            CRST(LRENT)~\cite{Zou_2019_ICCV} &42.7&47.0& +4.3\\
            CRST(MRKLD)~\cite{Zou_2019_ICCV} &43.8&47.3& +3.5\\
            \bottomrule
        \end{tabular}
        }
  }
 }
\captionsetup{font=footnotesize}
\caption{Performance improvements in mIoU of integrating our
TPLD with existing self-training adaptation approaches. We use the Deeplabv2-R segmentation model.}
\label{tab:self_apply}
\end{table*}
\subsection{Ablation study} 
\subsubsection{Lowering the selection threshold of CRST}
A straightforward way to generate dense pseudo labels is by lowering the selection threshold (i.e., increasing $p$) of self-training models.
We summarize the results in~\tabref{tab:frame}.
Since the scheme brings unconfident predictions at an early stage, either limited improvement ($p=0.4$, 47.0 $\to$ 47.1 mIoU) or worse performance is obtained ($p=0.6$, 47.0 $\to$ 45.7 mIoU).
Compared to these naive baselines, our TPLD shows significant improvement (47.0 $\to$ 51.2 mIoU).

\subsubsection{Framework design choices}
The main components of our framework design are the two-phase pseudo label densification.
The ablation results are shown in~\tabref{tab:frame}.
If we drop the voting stage, the model is trained alone with the easy-hard classification stage.
However, using full pseudo labels without any proper early-stage training introduces too noisy training signals (51.2 $\to$ 38.1 mIoU). If we drop the easy-hard classification stage, the model misses a chance to receive rich training signals from the full pseudo labels (51.2 $\to$ 49.5 mIoU).
We also explore the effect of ordering.
We observe that the voting-first method performs better than the easy-hard classification-first method (51.2 vs. 49.1 mIoU).
This implies that gradual densification is indeed important for stable model training.

\subsubsection{Effect of $\frac{1}{\lambda_{k}}$ in confidence score $conf_{\mathrm{t}}$}
We suggest to multiply $\frac{1}{\lambda_{k}}$ in computing the confidence score $conf_{\mathrm{t}}$.
The rationale behind this is to oversample the images, which include rare classes, and thus prevent the learning from being biased by images composed of obvious frequent classes. The results without and with the $\frac{1}{\lambda_{k}}$ are (50.5 vs 51.2 mIoU) and (33.7 vs 35.1 R-mIoU). This demonstrates the efficacy of incorporating $\frac{1}{\lambda_{k}}$.

\begin{table*}[t]
\setlength{\tabcolsep}{1pt}
\begin{subtable}{0.304\textwidth}{\renewcommand\arraystretch{0.91}
    \tiny
    \centering
\resizebox{\textwidth}{!}{
    \begin{tabular}{|c|c|cc|c|} 
    \hline
      Model & $p$ & Voting & EH Class. & \textbf{mIoU}\\
      \hline
        \multirow{3}{*}{CRST} & 0.2 &\multicolumn{2}{c|}{\xmark} & 47.0\\
        \cline{2-5}
        & 0.4 &\multicolumn{2}{c|}{\xmark} & 47.1\\
        \cline{2-5}
        & 0.6 &\multicolumn{2}{c|}{\xmark} & 45.7\\
        \hline
        \multirow{4}{*}{TPLD} & \multirow{4}{*}{0.2} & & \checkmark &38.1\\
        \cline{3-5}
        & & \checkmark & &49.5\\
        \cline{3-5}
        & & $\checkmark_{2}$ & $\checkmark_{1}$ &49.1\\
        \cline{3-5}
        & & $\checkmark_{1}$ & $\checkmark_{2}$ & \textbf{51.2}\\
        \hline
    \end{tabular}
    }
    }
   \caption{Framework design choices}
    \label{tab:frame}
\end{subtable}%
\hfill
\begin{subtable}{0.33\textwidth}{\renewcommand\arraystretch{1.12}
    \small
    \centering
\resizebox{\textwidth}{!}{
        \begin{tabular}[c]{|c | c | c | c|}
        \hline
        \fontsize{5.3}{5.3}\selectfont{\diagbox[width=10em]{Voting Num}{Voting Field}} &\fontsize{5.3}{5.3}\selectfont{37} & \fontsize{5.3}{5.3}\selectfont{\textbf{57}} & \fontsize{5.3}{5.3}\selectfont{77} \\
        \hline
        \fontsize{5.3}{5.3}\selectfont{1} & \fontsize{5.3}{5.3}\selectfont{48.61} & \fontsize{5.3}{5.3}\selectfont{48.95} & \fontsize{5.3}{5.3}\selectfont{48.57}\\
        \hline
        \fontsize{5.3}{5.3}\selectfont{\textbf{3}} & \fontsize{5.3}{5.3}\selectfont{49.49} & \fontsize{5.3}{5.3}\selectfont{\textbf{49.52}} & \fontsize{5.3}{5.3}\selectfont{48.37}\\
        \hline
         \fontsize{5.3}{5.3}\selectfont{5} & \fontsize{5.3}{5.3}\selectfont{48.72} & \fontsize{5.3}{5.3}\selectfont{48.00} & \fontsize{5.3}{5.3}\selectfont{48.63} \\
        \hline
        \end{tabular}
      }
      }
        \caption{Voting field / number}
    \label{tab:VF_VN}
\end{subtable}%
\hfill
\begin{subtable}{0.0999\textwidth}{\renewcommand\arraystretch{1.7}
    \tiny
    \centering
\resizebox{\textwidth}{!}{
    \begin{tabular}{|c|c|c|} 
    \hline
    $\alpha$ & \textbf{mIoU}\\
        \hline
         0.6 & 48.34 \\
        \hline
          \textbf{0.7} & \textbf{49.52} \\
          \hline
          0.8 & 49.25 \\
        \hline
    \end{tabular}
    }
    }
        \caption{$\alpha$}
    \label{tab:alpha}
\end{subtable}%
\hfill
\begin{subtable}{0.103\textwidth}{\renewcommand\arraystretch{0.95}
    \tiny
    \centering
\resizebox{\textwidth}{!}{
    \begin{tabular}{|c|c|} 
    \hline
      \textbf{$q$} & \textbf{mIoU}\\
      \hline
        1.00 &  48.44 \\
        \hline\hline
        0.40 &  50.15\\ 
        \hline
        0.35 &  50.27 \\
        \hline
        \textbf{0.30} & \textbf{51.20}\\
        \hline
        0.25 & 49.81 \\
        \hline
        0.20 & 50.02 \\
        \hline
    \end{tabular}
    }
    }
        \caption{$q$}
    \label{tab:que}
\end{subtable}%
\hfill
\begin{subtable}{0.0999\textwidth}{\renewcommand\arraystretch{1.7}
    \tiny
    \centering
\resizebox{\textwidth}{!}{
    \begin{tabular}{|c|c|c|} 
    \hline
    $\gamma$ & \textbf{mIoU}\\
        \hline
         1.5 & 50.0 \\
        \hline
          \textbf{2} & \textbf{51.2} \\
          \hline
          2.5 & 48.6 \\
        \hline
    \end{tabular}
    }
    }
        \caption{$\gamma$}
    \label{tab:gamma}
\end{subtable}%
\caption{\textbf{Results of ablation studies.}}
\label{tab:abla}
\end{table*}

\begin{table*}[t]
 \centering
  \setlength{\tabcolsep}{9pt}
  \subfloat[\scriptsize Exps on \nth{1} phase obj func.]{%
  {\renewcommand\arraystretch{1.1}
    \resizebox{0.36\textwidth}{!}{%
        \begin{tabular}{c c c c}
            \toprule
                                            & Bootstrap   & Voting      &\multirow{2}{*}{mIoU}\\
             & Eq.\eqref{eq:eq_5} & Eq.\eqref{eq:eq_4} & \\                                            
            \midrule
            $\mathcal{L}_{st}$~\cite{Zou_2019_ICCV} &            &              &47.00\\
                                                    & \checkmark &              &48.47\\
            $\mathcal{L}_{st1}$               & \checkmark & \checkmark   &49.52\\
            \bottomrule
        \end{tabular}
        }
  }
 }
\setlength{\tabcolsep}{9.5pt}
  \subfloat[\scriptsize Exps on \nth{2} phase obj func.]{%
  {\renewcommand\arraystretch{1.1}
    \resizebox{0.5\textwidth}{!}{%
        \begin{tabular}{c c c c}
            \toprule
                & EH Cls.   & Adv. &\multirow{2}{*}{mIoU}\\
                 & Eq.\eqref{eq:eq_7} + Eq.\eqref{eq:eq_8} & Eq.\eqref{eq:eq_9}+Eq.\eqref{eq:eq_10} & \\   
            \midrule
            $\mathcal{L}_{st1}$                     &            &              &49.51\\
                                                    & \checkmark &              &50.11\\
            $\mathcal{L}_{st1} + \mathcal{L}_{st2}$ & \checkmark & \checkmark   &51.20\\
            \bottomrule
        \end{tabular}
        }
  }
 }
\captionsetup{font=footnotesize}
\caption{\textbf{Detailed analysis on the proposed objective functions.} We note the corresponding equations for each proposals. \textit{Adv.} denotes adversarial loss term for hard sample training.}
\label{tab:loss_abl}
\end{table*}

\begin{figure*}[t]
    \centering 
    \includegraphics[width=270pt, height = 110pt]{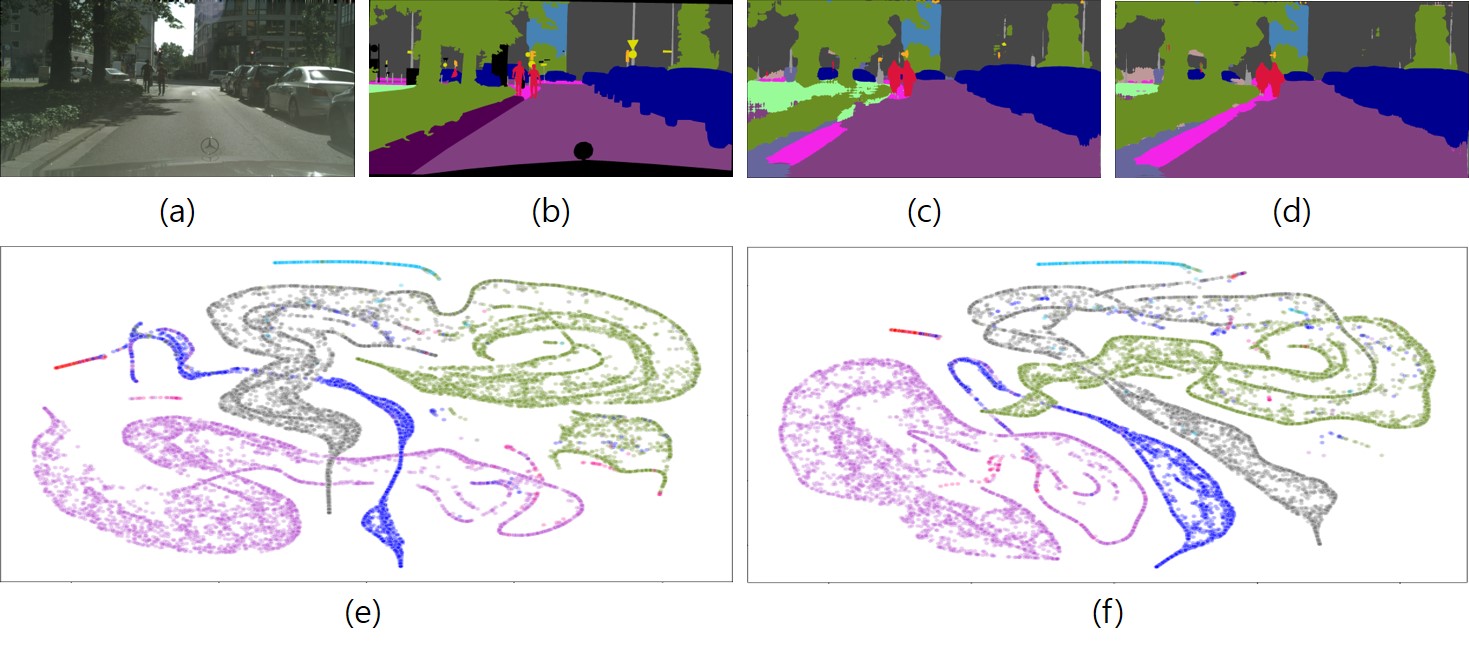}
    \caption{\textbf{A contrastive analysis of with and without hard sample training (Eq.\eqref{eq:eq_9}+Eq.\eqref{eq:eq_10}).} (a): target image, (b): ground truth, (c): prediction result without hard sample training, (d): prediction result with hard sample training. We map high-dimensional features of (c) and (d) to 2-D space features of (e) and (f) respectively using t-SNE~\cite{tsne}. 
    }
    \label{fig:tsne}
\end{figure*}

\subsection{Parameter analysis}
Here, we conduct experiments to decide optimal hyper-parameters in our framework.
For the first phase, we have a total of three hyper-parameters; voting field size, voting iteration number, and $\alpha$.
In ~\tabref{tab:VF_VN}, we conduct a grid search on the first two, and we obtain the best result with voting field 57, and voting number 3.
The hyperparameter $\alpha$ controls how much to maintain the initial prediction value, and we observe that 0.7 produces the best result (see~\tabref{tab:alpha}). 
We see that the results are in the same line with the residual learning~\cite{he2016deep}. 
Providing residual features (i.e., pooled neighboring confident prediction values) while securing the initial behavior (i.e., initial prediction values) is important.
For the second phase, we have a total of two hyper-parameters; $q$ and $\gamma$.
The hyperparameter q controls the `easy' portions in the target images.
For example, if we increase the value, more images will be used as easy samples for the training.
We observe that setting q to 0.3 provides the best result (see~\tabref{tab:que}).
Note that if we set q to 1 (i.e., making all the target images to be trained with the full pseudo labels), we instead obtain degraded performance. This implies that a proper portion of easy and hard samples are need to be set, and both the full pseudo label training and hard-to-easy feature alignment are important.
The hyperparameter $\gamma$ is related to the calibration degree of the prediction values in generating full pseudo labels (see Eq.~\eqref{eq:eq_7}). We obtain the best result when $\gamma$ equals 2.

\subsection{Loss function analysis}

Finally, we explore the impact of loss functions in \tabref{tab:loss_abl}.
We begin with the standard self-training loss, \textbf{$\mathcal{L}_{st}$}.
Introducing the bootstrapping mechanism boosts the performance significantly, from 47.00 to 48.47 mIoU.
This implies that explicitly handling noisy pseudo labels is crucial but lacking in the original formulation.
Also, using \textit{voting} to densify the sparse pseudo labels further pushes the performance from 48.47 to 49.52 mIoU.
The densified pseudo labels help model learning due to the increased training-signals and are complementary to the bootstrapping effect.
In the second phase, we investigate the impact of both easy sample training (EH Cls.) and hard sample training (Adv.).
The easy sample training pushes the performance from 49.52 to 50.11 mIoU, and the hard sample training further increases the performance from 50.11 to 51.20. The results demonstrate that the full-pseudo label training is indeed important and the hard-to-easy feature alignment further enhances the model learning. 
Especially for the hard sample training, we conduct a contrastive analysis in~\figref{fig:tsne}.
We observe that hard sample training improves category-level feature alignment (\figref{fig:tsne} (e)$\rightarrow$\figref{fig:tsne} (f)), and thus the prediction values become more accurate and clean (\figref{fig:tsne} (c)$\rightarrow$\figref{fig:tsne} (d)). 

\section{Conclusions}
In this paper, we point out that self-training methods for UDA suffer from the sparse pseudo label during training. Therefore, we present a novel two-phase pseudo label densification method.
Combined with recently proposed CRST framework, we achieve new state-of-the-art results on UDA benchmarks.\\

\noindent\textbf{Acknowledgement}
This research is supported by the National Cancer Center(NCC).

%
%
\bibliographystyle{splncs04}
\bibliography{eccv2020_camready}
\end{document}